\documentclass[11pt]{article}
\usepackage[preprint]{acl}

\usepackage{multirow}
\usepackage{booktabs} % For \toprule, \midrule, \bottomrule
\usepackage{amsmath}  % For \vec and mathematical symbols
\usepackage{float}
\usepackage{amsfonts}

\usepackage{times}
\usepackage{latexsym}
\usepackage{hyperref}

\usepackage[T1]{fontenc}

\usepackage[utf8]{inputenc}

\usepackage{microtype}

\usepackage{inconsolata}

\usepackage{graphicx}

\title{Supervised Semantic Differential for Cross-Cultural Concept Analysis: \\
A Case Study of Human Affect}

\author{Jan Sikora \\
  University of Warsaw \\
  \texttt{jansikora.pub@gmail.com} \\\And
  Paweł Lenartowicz \\
  Society for Open Science \\
  Centre for Brain Research, \\
  Jagiellonian University \\
  \texttt{pawellenartowicz@europe.com} \\\And
  Hubert Plisiecki \\
  IDEAS Research Institute \\
  \texttt{hplisiecki@gmail.com} \\
}

\begin{document}
\maketitle
\begin{abstract}
Cross-cultural comparison of psychological meaning requires methods that
go beyond word-level translation and examine how semantic dimensions are
organized across languages. We introduce a cross-lingual extension of the
Supervised Semantic Differential (SSD), which estimates supervised
semantic gradients in embedding space and compares them across aligned
multilingual word embeddings. The method tests gradient alignment and
difference using permutation procedures and bootstrap intervals, and
interprets residual differences through clustering around the difference
gradient. We demonstrate the approach on Polish, English, and French
affective norm lexicons, modeling Valence, Arousal, and Dominance where
available. Affective dimensions were significantly recoverable across
languages and model settings. Cross-lingual comparisons showed broad
alignment together with structured residual differences: Valence appeared
mostly shared, whereas Arousal and Dominance produced more interpretable
contrasts involving bodily threat, aesthetic stimulation, internal
emotionality, macro-level authority, and everyday control. Several
clusters also reflected corpus-specific artifacts, underscoring the need
for cautious interpretation. Cross-lingual SSD offers an explainable
framework for testing semantic alignment, identifying divergence, and
generating hypotheses about cross-cultural differences in psychological
meaning. 
Code to reproduce our results is available at \url{https://github.com/przebor/Cross-Cultural-SSD}.
\end{abstract}

\section{Introduction}

Cross-cultural psychology studies how concepts vary
across cultures and languages. Cultural psychology has long argued that
psychological processes such as cognition, emotion, and motivation are
partly organized through culturally patterned models of selfhood and
social life \citep{markus_culture_1991}. At the same time, linguistic
approaches to culture emphasize that concepts do not always map cleanly
across languages, because lexical meanings are embedded in
culture-specific systems of interpretation \citep{wierzbicka_emotions_1999}.
This problem is especially important for psychological constructs, where
the same label may not refer to exactly the same conceptual structure
across cultural contexts. Cross-cultural comparison therefore requires
methods that can move beyond asking whether individual words can be
translated, and instead ask how larger regions of meaning are organized
across languages.

NLP research offers new tools for this problem. Recent work on
cross-cultural NLP argues that cultural variation is not reducible to
language coverage alone: speakers may differ in common ground,
communicative norms, values, and the kinds of meanings they encode in
text \citep{hershcovich_challenges_2022}. Broader surveys of culturally
aware NLP likewise emphasize the need for methods that can represent,
adapt to, and evaluate culturally situated meaning
\citep{liu_culturally_2025}. Word embedding models provide one route
toward this goal by representing lexical meaning geometrically, making it
possible to study concepts not only as isolated words, but as directions,
neighborhoods, and regions in semantic space. Early work on word
embeddings showed that semantic and syntactic regularities can often be
captured through vector-space relations
\citep{mikolov_linguistic_2013}, and related work in computational
cultural analysis has used such geometric structure to study cultural
associations and categories \citep{kozlowski_geometry_2019}.
Cross-lingual embedding methods extend this idea by aligning
independently trained monolingual spaces into a shared multilingual
space, often through orthogonal transformations
\citep{conneau2017word}. These models make it possible to compare
semantic structure across languages, creating an opportunity for
computationally explicit cross-cultural concept analysis.

In this paper, we focus on the cross-cultural analysis of affective
meaning. Emotions are a particularly important test case because they
have long been central to debates about universality and cultural
specificity. Classic work on facial expressions argued for pan-cultural
associations between facial movements and discrete emotion categories
\citep{ekman_pan-cultural_1969}. In contrast, constructionist and
cultural approaches argue that emotions are shaped by learned concepts,
language, bodily inference, and culturally patterned forms of social life
\citep{barrett_theory_2016,mesquita_cultural_2016}. Empirical work with
remote cultural groups has likewise challenged strong claims of
universality in emotion perception, suggesting that the interpretation of
emotional expressions depends on culturally available concepts and task
contexts \citep{gendron_perceptions_2014}. These debates make affective
meaning a useful domain for testing whether multilingual semantic models
can reveal both shared and language-specific conceptual structure.

Affective meaning is often studied through broad dimensions rather than
through discrete emotion categories alone. The semantic differential
tradition proposed that connotative meaning is organized around
Evaluation, Potency, and Activity \citep{osgood_measurement_1978}, and
cross-cultural work suggested that these dimensions recur across diverse
linguistic communities \citep{osgood_cross-cultural_1975}. Circumplex
models similarly describe affective meaning in terms of dimensions such
as Valence and Arousal, with evidence that such structures are at least
partially recoverable across languages, including Polish and English
\citep{russell_cross-cultural_1989}. However, the presence of shared
affective dimensions does not imply that the same words, semantic
neighborhoods, or conceptual domains define those dimensions in every
language. Large-scale lexical norming datasets make this question
empirically tractable: \citet{warriner_norms_2013} provide ratings of
Valence, Arousal, and Dominance for 13,915 English lemmas, while
\citet{imbir_affective_2016} provide affective and psycholinguistic
ratings for 4,905 Polish words and \citet{monnier_affective_2014} provide ratings of Valence and Arousal for 1,031 French words. These resources allow affective meaning
to be modeled not only as individual word-level ratings, but as
systematic dimensions of semantic variation.

To better understand how psychological phenomena differ across culture, we extend the Supervised Semantic Differential (SSD) for comparing affective dimensions across languages.
SSD estimates a semantic gradient in an embedding space by relating word
or text representations to an external psychological variable, and then
interprets the resulting direction through neighboring words, clusters,
or text examples \citep{plisiecki_measuring_2026}. Here, we focus on a specific extension
of SSD which uses a partial least squares (PLS) to fit the gradient and tests the fitted
gradient for significance with a corrected resampled $t$-test over
repeated train--test splits
\citep{nadeau_inference_2003, anonymous_cheap_2026}. We start by estimating
Valence, Arousal, and Dominance gradients for Polish and English and the Valence and Arousal gradients for French. We
first fit these gradients in monolingual embedding spaces to test whether
affective dimensions are recoverable within each language. We then fit
the same gradients in a shared multilingual embedding space, allowing the
resulting Polish, English and French affective directions to be pairwise compared
directly with regards to whether they differ in systematic ways.

Our contribution is a method for statistically testing and interpreting
cross-lingual differences in semantic dimensions. Rather than comparing
individual translated words, we compare whole affective gradients in a
shared embedding space. We use permutation tests to determine whether gradients differ across languages more than expected by chance, and we
interpret the resulting difference gradients through clustering in
semantic space. This provides an explainable framework for studying how
psychological dimensions of meaning are preserved, shifted, or
reorganized across languages. In doing so, the paper connects
cross-cultural psychology, the semantic differential tradition, and
modern multilingual representation learning, providing a computational
method for cross-cultural concept analysis.

\section{Methodology}
\label{sec:methodology}
\subsection{Problem statement}
Supervised Semantic Differential allows for study of the relation of an external variable to language usage, but can't be utilized by default for cross-cultural (cross-lingual) difference analysis, as the traditional word embedding models for different languages were not trained to be aligned and thus, are not -- beyond standard chance.

In order to compare psychological variables across texts in different languages and cultures, we propose using multilingual word embedding models, together with permutation tests to assess the significance of the gradient difference and a bootstrap to quantify its stability, and applying k-means silhouette clustering of words from the word embeddings models around the difference gradient, as in the SSD method \citep{plisiecki_measuring_2026}.

\subsection{Multilingual word embeddings}
The multilingual word embedding models are separate word embedding models for each language, aligned in such a way that allows for cross-model comparisons.
We propose using multilingual models that went through supervised training on word translation pairs so that the embeddings of the same word in both models are possibly close. They are typically trained by fitting a linear transformation matrix between existing, pretrained word embeddings in both languages \citep{mikolov_exploiting_2013}.
It has also been shown that normalizing the vector embeddings of the words and consequently applying a constraint for orthogonality of the linear transformation matrix improves the mapping performance \citep{xing_normalized_2015}.

\subsection{SSD fit}
\label{subsec:fit}
We propose using the Supervised Semantic Differential method with a partial least squares (PLS) backend \citep{anonymous_cheap_2026} to fit separate gradients that represent the studied independent variable for both languages using two different models, thus resulting in two gradients -- $g_A$ and $g_B$. The number of retained PLS components $K$ is chosen by the one-standard-error rule \citep{breiman_classification_1984}, and the significance of each fitted gradient is assessed with a corrected resampled $t$-test over repeated train--test splits.

\subsection{Significance and Robustness Analysis}
\label{subsec:significance}

The significance of each fitted gradient is established by the corrected resampled $t$-test of Section~\ref{subsec:fit}. To compare $g_A$ and $g_B$ across languages we run two permutation tests and a bootstrap: one test asks whether the gradients are aligned beyond chance, the other whether they differ, and the bootstrap measures how stable that comparison is.

\subsubsection{Test statistic}
Both tests use the cosine similarity between the two gradients,
\begin{equation}
  \rho \;=\; \cos(g_A, g_B) \;=\; g_A^\top g_B,
\end{equation}
which equals the inner product because the gradients are unit-normalized. For
robustness and reproducibility, the gradients are recomputed during permutation with
the closed-form single-component PLS1 solution ($K=1$). For embeddings standardized
column-wise into $Z$, with $Z_{ij}=(x_{ij}-\bar{x}_j)/s_j$ and $s_j$ the
standard deviation of coordinate $j$, and z-scored labels $\tilde{y}$,
\begin{equation}
  g \;=\; \frac{\beta}{\lVert\beta\rVert_2}, \qquad
  \beta_j \;=\; \frac{1}{s_j}\,\bigl(Z^\top \tilde{y}\bigr)_j .
\end{equation}
At $K=1$ the gradient orientation is fixed, so $\rho$ is reproducible.

\subsubsection{Alignment test (\texorpdfstring{$H_0:\rho=0$}{H0:rho=0})}
Does the alignment exceed chance? Keeping each embedding matrix fixed, we
shuffle the word-to-label pairing within each dataset, refit both gradients,
and recompute $\rho$. Shuffling removes any word--label link, so the two
refitted gradients are independent and $\mathbb{E}[\rho]=0$. Repeating this $n$
times gives the null, and the upper-tail $p$-value (with a pseudocount so it is
never zero) is
\begin{equation}
  p_{\text{align}} \;=\; \frac{1+\sum_{t=1}^{n}\bigl[\rho^{(t)}\ge\rho\bigr]}{n+1},
\end{equation}
where $\rho^{(t)}$ is the cosine at permutation $t$. A small $p_{\text{align}}$
means the two languages recover the same affective dimension.

\subsubsection{Difference test (\texorpdfstring{$H_0:\rho=1$}{H0:rho=1})}
Do the gradients differ? We pool both lexicons (labels z-scored within
language), shuffle the language label while keeping the group sizes fixed,
refit one gradient per group, and recompute $\rho$. Under exchangeability the
two groups are random halves of one pool, so their gradients nearly coincide
and the null $\rho$ concentrates near $1$; a real difference pushes the observed
$\rho$ into the lower tail. The $p$-value $p_{\text{diff}}$ uses the same
formula with $\ge$ replaced by $\le$. A small $p_{\text{diff}}$ means the
gradients differ across languages. Because the test uses each full lexicon,
exchangeability is only approximate: a low $p_{\text{diff}}$ may reflect a real
gradient difference or differences in lexical composition (see the Limitations Section).

\subsubsection{Bootstrap interval}
The bootstrap gauges how precisely $\rho$ is estimated. Resampling
\textit{(word, label)} pairs with replacement within each lexicon, we refit both
gradients and recompute $\rho$; the scatter of these resampled cosines measures
its sampling error. We use a \emph{hybrid} interval: only the spread is borrowed
from the bootstrap, while the interval stays centered on the observed $\rho$
rather than the bootstrap mean, which is biased downward by resampling noise.
The spread is measured on the Fisher-$z$ scale $z=\operatorname{arctanh}(\rho)$,
giving a standard error $\hat{\sigma}_z$, and the $95\%$ interval is
back-transformed,
\begin{equation}
  \bigl[\tanh(z_{\text{obs}} - 1.96\,\hat{\sigma}_z),\;
        \tanh(z_{\text{obs}} + 1.96\,\hat{\sigma}_z)\bigr],
\end{equation}
where $z_{\text{obs}}=\operatorname{arctanh}(\rho)$. The transform keeps the
interval from crossing the $\rho=1$ boundary.

\subsection{Difference gradient analysis and clustering}
If the significance analysis reveals that the gradient difference is statistically significant, we propose treating the resulting $\Delta{\vec{g}}=\vec{g}_A-\vec{g}_{B}$ gradient as if it was the result of the traditional Supervised Semantic Differential method - finding word clusters, by considering embeddings of the vocabularies of the models and using $k$-means clustering around $\Delta{\vec{g}}$ and the opposite $-\Delta{\vec{g}}$, with $k$ chosen to maximize the silhouette score \citep{plisiecki_measuring_2026}. The resulting clusters should represent concepts that cause the underyling cross-cultural difference in the psychological variable.

\begin{table*}[t] % [t] is much more stable for double-column tables
  \centering
  \begin{tabular}{llrlrcrcc}
    \toprule
    \textbf{Dataset} & \textbf{Model} & \textbf{$N$} & \textbf{Dimension} & \textbf{$K$} & \textbf{$R^2$} & \textbf{$p$-val} & \textbf{$r_{pred}$} \\ \midrule
    % Added {6} to multirow so it knows how many rows to span
    \multirow{6}{*}{IMBIR (PL)} & \multirow{3}{*}{Multilingual} & \multirow{3}{*}{3584} 
     & Valence & 3 & 0.578 & $^{***}$ & 0.736 \\
     & & & Arousal & 3 & 0.542 & $^{***}$ & 0.669 \\
     & & & Dominance & 3 & 0.463 & $^{***}$ & 0.644 \\
     \cmidrule(lr){2-8}
     & \multirow{3}{*}{Monolingual} & \multirow{3}{*}{4656} & Valence & 3 & 0.662 & $^{***}$ & 0.810 \\
     & & & Arousal & 3 & 0.590 & $^{***}$ & 0.663 \\
     & & & Dominance & 3 & 0.576 & $^{***}$ & 0.751 \\ \midrule
    \multirow{6}{*}{Warriner (EN)} & \multirow{3}{*}{Multilingual} & \multirow{3}{*}{13541} & Valence & 4 & 0.584 & $^{***}$ & 0.731 \\
     & & & Arousal & 4 & 0.360 & $^{***}$ & 0.553 \\
     & & & Dominance & 4 & 0.459 & $^{***}$ & 0.646 \\
     \cmidrule(lr){2-8}
     & \multirow{3}{*}{Monolingual} & \multirow{3}{*}{13817} & Valence & 5 & 0.627 & $^{***}$ & 0.770 \\
     & & & Arousal & 4 & 0.360 & $^{***}$ & 0.581 \\
     & & & Dominance & 4 & 0.474 & $^{***}$ & 0.672 \\ \midrule
         \multirow{4}{*}{FAN (FR)} & \multirow{2}{*}{Multilingual} & \multirow{2}{*}{993} & Valence  & 2 & 0.588 & $^{***}$ & 0.731 \\
     & & & Arousal & 2 & 0.516 & $^{***}$ & 0.553 \\
     \cmidrule(lr){2-8}
     & \multirow{2}{*}{Monolingual} & \multirow{2}{*}{994} & Valence & 2 & 0.635 & $^{***}$ & 0.770 \\
     & & & Arousal & 3 & 0.656 & $^{***}$ & 0.581 \\ \bottomrule
     \multicolumn{8}{l}{$^{***}$ denotes $p < 0.001$}
  \end{tabular}
  \caption{SSD Fitting results for Valence, Arousal, and Dominance. $N$ represents the number of words in the lexicon and the word embedding model, $K$ the number of retained PLS components, and $r_{pred}$ the correlation between predicted and actual values.}
  \label{tab:ssd-results} % Always put the label AFTER the caption
\end{table*}

\section{Case study: affective norms across cultures}

\subsection{Lexicons for modeling affective norms}
To test the performance of the introduced method, we utilize three datasets of Polish, English and French word lexicons with annotated numerical scores representing the Valence, Arousal and Dominance affect labels of words in Polish and English and of Valence and Arousal labels in French. 
The Polish lexicon comes from the dataset by \citet{imbir_affective_2016} ($n=4905$), the English lexicon from \citet{warriner_norms_2013} work ($n=13915$) and the French lexicon from the \citet{monnier_affective_2014} dataset ($n=1031$).
We first preprocess the data by lemmatizing the words and then removing any stopwords using the models: \textit{pl\_core\_news\_lg} for Polish, \textit{en\_core\_web\_lg} for English, and \textit{fr\_core\_news\_lg} for French, from the  \textit{spacy} package \citep{Honnibal_spaCy_Industrial-strength_Natural_2020}.

\subsection{SSD on monolingual models}
\label{subsec:monolingual}
At this stage, we additionally utilize the Supervised Semantic Differential method \citep{plisiecki_measuring_2026} with a PLS backend \citep{anonymous_cheap_2026} and monolingual word embedding models as a baseline to show that every affective dimension constitutes recoverable linguistic patterns and that the traditional SSD mono-lingual analysis is able to recover them.

We fit separate VAD gradients for Polish, English and French and observe whether the results are statistically significant. The utilized English word embedding model is the 2024 Dolma release of GloVe \citep{carlson_new_2025} with 300 dimensions and the vocabulary size of 1.2 million, French model is the word2vec model from \citet{fauconnier_2015} with 500 dimension and the Polish model is the GloVe model from \citet{dadas_2019} with 300 dimensions. Before analysis, we L2 normalize the models' vectors and remove the first component of variance to reduce anisotropy, as per the original implementation \citep{plisiecki_measuring_2026}.

\subsection{Multilingual analysis using the Methodology}

To overcome the gradient comparability constraint, we utilize the, previously introduced in Section \ref{sec:methodology}, Methodology for cross-cultural analysis. We use separate Polish, English and French word embeddings models aligned in the same vector space using, as discussed, orthogonal linear transformations. The models come from \citeposs{conneau2017word} work. We apply the same L2 normalization, and first component of variance removal, as in the case of the monolingual models \citep{plisiecki_measuring_2026}.
 
We fit the new SSD gradients on these multilingual models and expect gradient divergence reflecting the underlying cultural differences. To properly determine whether the observed difference between SSD gradients is significant, we apply the previously introduced permutation tests and the bootstrap providing confidence intervals on the difference.

If the mentioned tests reveal significant results, we use the difference vector $\Delta \vec{g} = \vec{g}_{a} - \vec{g}_{b}$ for all language pairs $(a,b)\in\{(pl, en), (en, fr), (fr, pl)\}$ as if it was a result of the traditional SSD fit and use the k-means silhouette clustering to reveal word groups that describe what exactly causes affective norm differences between languages. For the clustering, we consider vocabularies from both language embedding models separately.

\section{Case Study Results}
\subsection{Statistical Significance of the Method}
\label{subsec:results-permutation}
The corrected resampled $t$-tests confirm that each fitted gradient is statistically significant (Table~\ref{tab:ssd-results}) -- the affective norm dimensions are successfully captured with high statistical significance ($p < 0.001$) across all languages and both model types.

The permutation tests (with $n=1000$) against exchangeable labels reject the null of no cross-lingual alignment for every language pair and affect dimension ($p<0.001$), indicating that the affect scores constitute a cross-cultural measure of the emotional understanding of words. The bootstrap procedure yields the following 95\% confidence intervals for the difference between language gradients:

\begin{figure*}[t]
  \includegraphics[width=1.0\linewidth]{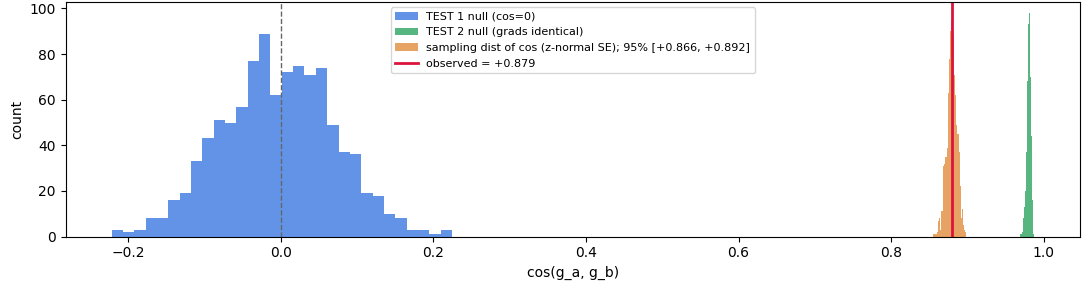}
  \caption{Permutation tests null distributions and sampling error distributions for the Valence dimension of Polish - English pair.}
  \label{fig:valence-permutation}
\end{figure*}

\begin{table}[H]
    \centering
    \small

    \begin{tabular}{lccc}
        \toprule
        \textbf{Pair} & \textbf{Valence} & \textbf{Arousal} & \textbf{Dominance} \\
        \midrule
        EN -- PL & $[0.87, 0.89]$ & $[0.79, 0.82]$ & $[0.64, 0.72]$ \\
        EN -- FR & $[0.69, 0.80]$ & $[0.64, 0.73]$ & N/A \\
        PL -- FR  & $[0.64, 0.75]$ & $[0.65, 0.74]$ & N/A \\
        \bottomrule
    \end{tabular}
    \caption{Bootstrapped 95\% Confidence Intervals for Cross-Lingual Gradient Difference}
    \label{tab:ssd-cis}
\end{table}

% Detailed look at the specific words/clusters found in 3.3. 
% For example, identifying if "Arousal" in Polish is more associated with 
% negative stress vs. English "Arousal" associated with excitement.
\subsection{Difference-Gradient Word Clusters}
Analysis of the clustering done over the $\Delta \vec{g}$ and $-\Delta \vec{g}$ reveals distinct clusters of words highlighting specific semantic domains where English, Polish and French speakers differ most in their conceptualization of Valence, Arousal, or Dominance. In this section, we will attempt to qualitatively underline the most important differences in human affect across these three cultures. All of the examples mentioned in the following sections were translated to English, and selected from the full cluster results (available in the Appendix \ref{sec:full-clusters}).

\subsubsection{Internal-External Arousal}
A number of clusters related to the contrast between deeper, more abstract feelings and strictly physical or physiological matters emerged in pairs involving the French language and the Arousal dimension. Particularly, on the French side, there were clusters involving kinship (e.g. \textit{father, mother, siblings}) and profound emotional states (e.g. \textit{longing, yearning, sorrow}) in contrast to clusters on the English side, which heavily featured clinical sexual anatomy (e.g. \textit{vagina, penis, genital, ejaculation}) and physical geometry (e.g. \textit{cylindrical, angled, conical}).

This pattern of prioritizing the internal and prosocial over the physical was also present in the French-Polish comparison. While Polish divergence was heavily anchored in visceral bodily trauma (e.g. \textit{damage, bleeding}) and violent conflict (e.g. \textit{murdered, execution by shooting}), the French Arousal space was uniquely driven by prosocial joy (e.g. \textit{kindness, joy/delight}) and aesthetic leisure (e.g. \textit{landscapes, travel}). 

A difference can also be observed when looking at Arousal in the English -- Polish pair. The English understanding diverges toward external, structured stimulation, characterized by clusters emphasizing leisure (e.g. \textit{ecotourism, sightseeing}) and visual aesthetics (e.g. \textit{colorful, vibrant}). The Polish Arousal, however, remains strictly anchored in survival and severe physical peril. Its divergent clusters are dominated by violent acts (e.g. \textit{killed, murdered}) and debilitating physical conditions (e.g. \textit{chronic, hospitalization}). 

Consequently, the cross-cultural Arousal analysis shows a clear difference: French emotionality focuses on internal aspects, while English on aesthetic and leisure stimulation, and Polish on the existential threat.

\subsubsection{Macro-Micro Dominance}
Because of the lack of annotation of the Dominance affect (which measures feelings of agency, autonomy, and environmental control) in the French dataset, we were only able to qualitatively compare it in the English -- Polish pair. 

The comparison revealed a striking macro-versus-micro split between the Polish and English understanding. The Polish space diverged heavily toward themes of macro-power, and absolute top-down authority, featuring clusters centered on monarchy and rule (e.g. \textit{ruler, monarch}), structural violence and terrorism (e.g. \textit{terrorist, warlords}), and supernatural forces (e.g. \textit{demonic, monster}). Conversely, the English space diverged toward micro-control and personal environmental control. Its clusters were uniquely anchored in the mundane domestic sphere (e.g. \textit{kitchen, toilets, meals, beds}) and administrative tasks (e.g. \textit{apologies, report}). This suggests that while the Polish conceptualization of Dominance is uniquely shaped by forces of sweeping power, the English one locates feelings of control and autonomy within the everyday life of personal boundaries, daily routines, and bureaucratic order.

\subsubsection{Valence as a Mostly Shared Evaluative Axis}

Compared with Arousal and Dominance, Valence produced weaker and less
clearly interpretable difference gradients. This is consistent with the
high cross-lingual similarity of the Valence directions, especially in
the English -- Polish comparison, suggesting that positive--negative
evaluation is largely shared across the analyzed languages.

The residual clusters nevertheless revealed some local differences. In
the English -- Polish pair, the Polish-leaning pole contained terms related
to talent, esteem, education, creativity, and cultural description,
whereas the English-leaning pole was associated with accusation,
suspicion, blocking, confirmation, causation, and consequences. This may
suggest a weak contrast between valued personal or cultural
accomplishment and more legal, evidential, or causal forms of negative
evaluation.

The French comparisons were less interpretable. Polish -- French and
English -- French Valence clusters contained many botanical, technical,
geographic, and proper-name artifacts, alongside some socially negative
material (e.g. \textit{accusation, violation, homicide, arrogance,
pretentiousness, and unhappiness}). We therefore treat the Valence results
primarily as evidence for broad cross-lingual alignment, with only
limited exploratory evidence for language-specific semantic departures.

\subsubsection{Culture-Specific Linguistic Artifacts}
It is important to note that several divergent clusters, represent dataset- or model-driven linguistic artifacts rather than actual psychological shifts in affective processing. Across the language pairs, there were frequent isolated clusters composed entirely of highly localized proper nouns and toponyms. For example, the Polish space consistently clustered local municipalities and regions (e.g. \textit{krakowskich, wrocławskich, tarnów, ostrowiec}), while the English space highlighted North American geographic features and counties (e.g. \textit{allegheny, shenandoah, kanawha}). Similarly, the French space produced distinct, tightly coherent clusters of Québécois surnames (e.g. \textit{tremblay, gagnon, lévesque, falardeau}). Rather than reflecting a true cross-cultural misalignment in emotional evaluation, the emergence of these clusters likely stems from the geographic and cultural biases inherent in the text corpora used to train the respective embedding models. These localized entities carry dense contextual associations in their native language but can only be treated as infrequent, emotionally neutral loanwords in the contrasting languages.

\section{Discussion and Conclusion}

We introduced a cross-lingual extension of the Supervised Semantic
Differential for comparing psychological dimensions of meaning across
languages. Rather than comparing individual translated words, the method
fits supervised semantic gradients separately within each language and
then compares the resulting directions in a shared multilingual embedding
space. This makes it possible to ask whether a psychological dimension is
both recoverable within languages and organized similarly across them.
In the affective norm case study, Valence, Arousal, and Dominance were
all significantly recoverable in both monolingual and multilingual
settings, suggesting that affective ratings correspond to robust
directions in lexical semantic space.

The cross-lingual analyses showed a combination of broad alignment and
structured divergence. Valence appeared to be the most stable affective
dimension: the corresponding gradients were highly aligned, and the
difference-gradient clusters were comparatively weak and artifact-heavy.
This supports the interpretation of Valence as a largely shared
evaluative axis, with only limited language-specific residual structure,
consistent with the broader semantic differential tradition
\citep{osgood_cross-cultural_1975,osgood_measurement_1978}. Arousal
showed clearer semantic divergence. Across language pairs, the difference
clusters suggested contrasts between bodily threat, physiological
disruption, aesthetic stimulation, leisure, kinship, and internal
emotional experience. Dominance, available only for the English -- Polish
comparison, showed the clearest substantive contrast: Polish-leaning
clusters emphasized macro-level power, violence, and authority, whereas
English-leaning clusters emphasized domestic, bureaucratic, and everyday
forms of control.

These findings should be interpreted as exploratory evidence about
cross-lingual organization of affective meaning, not as direct evidence
that cultures differ in exactly the same way. The method detects
differences between supervised gradients estimated in aligned embedding
spaces. Such differences may reflect culturally patterned meanings, but
they may also reflect differences in lexicon composition, corpus domain,
register, frequency, and the quality of cross-lingual alignment
\citep{conneau2017word,xing_normalized_2015}. This was visible in the
cluster analyses, where several coherent clusters consisted primarily of
place names, surnames, technical terms, or other corpus-specific lexical
material. These artifacts are not merely noise: they show that
difference-gradient interpretation can reveal both substantive semantic
contrasts and the limits of the underlying representational resources.

An important use of the proposed framework is hypothesis generation. The
difference-gradient clusters do not by themselves establish the cultural
or psychological source of a cross-lingual difference, but they make such
differences empirically tractable by translating an abstract geometric
contrast into specific semantic domains. This is especially relevant for
Arousal. Prior work on affect valuation suggests that cultures differ not
only in whether affective states are positive or negative, but also in
which levels and forms of arousal are valued, promoted, and pursued
\citep{tsai_ideal_2007,lim_cultural_2016}. Our results do not test
ideal affect directly, but they suggest concrete hypotheses about the
semantic organization of arousal: whether Polish affective meaning gives
greater weight to bodily threat and physical harm, whether English
Arousal is more strongly associated with external stimulation and
aesthetic activity, and whether French Arousal is more closely tied to
interpersonal, autobiographical, or inwardly experienced emotion. The
Dominance results similarly suggest a testable contrast between
macro-level authority and threat on the Polish side and everyday,
domestic, or bureaucratic control on the English side. These hypotheses
can then be evaluated in follow-up studies using matched stimuli,
bilingual annotation, controlled corpora, or experimental rating tasks.
In this sense, cross-lingual SSD is not intended to replace direct
cross-cultural validation, but to guide it by identifying where such
validation should be most informative \citep{markus_culture_1991,
wierzbicka_emotions_1999}.

The main contribution of the paper is therefore methodological. The
proposed framework provides a way to test whether psychological
dimensions are aligned across languages, whether residual differences are
larger than expected under permutation, and which semantic neighborhoods
contribute most strongly to those differences. Importantly, the approach
does not require one-to-one translation pairs at the interpretation
stage. This makes it applicable to more complex cross-cultural datasets,
including open-ended responses, survey texts, interviews, or other
settings where psychological variables are available but direct lexical
translation is incomplete or theoretically insufficient
\citep{hershcovich_challenges_2022,liu_culturally_2025}.

Future work should validate the resulting difference gradients against
human judgments, matched bilingual lexicons, and multiple independently
trained multilingual embedding models. A particularly important next step
is to separate cultural effects from corpus and model effects by
replicating the analysis across balanced corpora and alternative
alignment methods. More generally, cross-lingual SSD can be extended
beyond affect to dimensions such as agency, morality, sociality,
certainty, ideology, or self-construal. Used in this way, the method
offers an explainable bridge between cross-cultural psychology and
multilingual representation learning: it treats psychological constructs
as semantic directions whose alignment, divergence, and interpretability
can be directly examined \citep{plisiecki_measuring_2026}.

\section*{Limitations}
\label{sec:limitations}
The present analysis relies on the assumption that affective and semantic
contrasts can be represented, at least approximately, as directions in
distributional embedding spaces. This assumption is related to prior work
showing that some linguistic relations are preserved as linear offsets in word
embeddings \citep{mikolov_linguistic_2013}, but it should not be understood as
a claim that all relevant cross-linguistic meaning differences are strictly
linear. The recovered gradients are therefore best interpreted as compact
approximations of broad affective contrasts, rather than as exhaustive
representations of the full semantic structure of each language.

A second limitation concerns multilingual alignment. The analysis assumes that
the multilingual embedding space places translation-equivalent and
semantically comparable words in sufficiently aligned regions of the shared
space. The fact that the difference-based procedure recovered statistically
distinct and interpretable clusters for the affective norms provides an
important internal check on this assumption. However, it does not guarantee
perfect alignment across languages. Residual misalignment may affect both the
estimated gradient directions and the nearest-neighbor vocabularies used for
interpretation, especially for culturally specific terms, idioms, or words
without close translation equivalents.

Finally, observed differences between datasets may partly reflect the
composition of the corpora used to train the underlying embedding models rather
than genuine cross-cultural differences in affective meaning. Differences in
domain, register, genre, time period, and topical coverage can induce
differences in embedding geometry that are not attributable to language or
culture as such. For this reason, the results should be interpreted as
differences between affective norms as represented in these particular
embedding spaces, not as direct evidence of stable cross-cultural differences
in emotional experience. Future work should test the theoretical conclusions of this analysis, using other cross-cultural methods to confirm their robustness.

The text of this manuscript was partially polished
with the assistance of a Large Language Model; all
revised passages were reviewed and corrected by
the author.

\section*{Ethical Considerations}

The present study analyzes aggregate semantic patterns in affective word norms
and embedding spaces, rather than individual-level data. The method is intended
to approximate cultural differences in affective meaning at the level of
lexical-semantic systems, not to profile, classify, or evaluate individual
speakers. Its results should therefore be interpreted as population- and
language-level semantic patterns, mediated by the specific norm datasets,
embedding models, and alignment procedures used in the analysis.

A further ethical concern is that pretrained embeddings may encode social,
cultural, and historical biases present in their training corpora. Such biases
are not separate from cultural meaning, but they can distort its measurement by
over-representing particular registers, domains, social groups, or historical
contexts. For this reason, the results should be used as evidence for
interpretable cross-cultural approximations of affective meaning, rather than
as rankings or evaluations of languages, cultures, or communities.

\bibliography{custom}

\appendix
\section{Significance and Robustness Test Results}
In this appendix, we present figures showcasing all the performed permutation tests and bootstraps, performed for every language pair and dimension (with the number of permutations $n=1000$), as established in Section \ref{subsec:significance}. The high-level overview of the results is also available in Section \ref{subsec:results-permutation}.

\subsection{English -- Polish}

\subsubsection{Valence}

\begin{figure}[H]
  \includegraphics[width=\columnwidth]{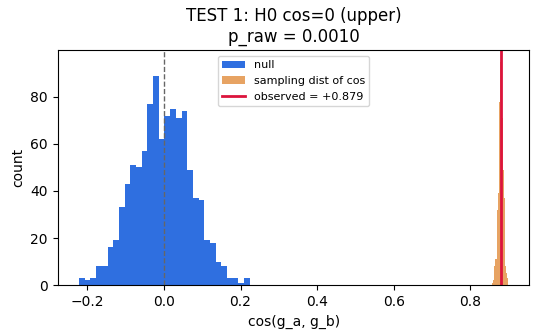}
  \caption{cos=0 test distribution for the Valence dimension of Polish - English pair.}
\end{figure}

\begin{figure}[H]
  \includegraphics[width=\columnwidth]{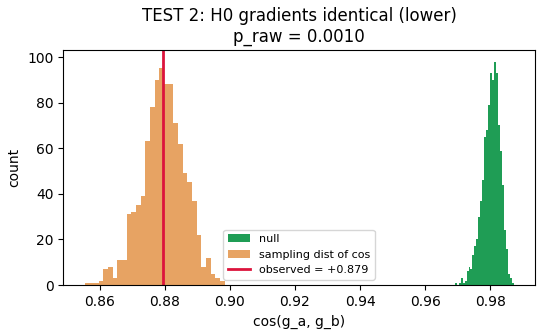}
  \caption{cos=1 test distribution for the Valence dimension of Polish - English pair.}
\end{figure}

\subsubsection{Arousal}

\begin{figure}[H]
  \includegraphics[width=\columnwidth]{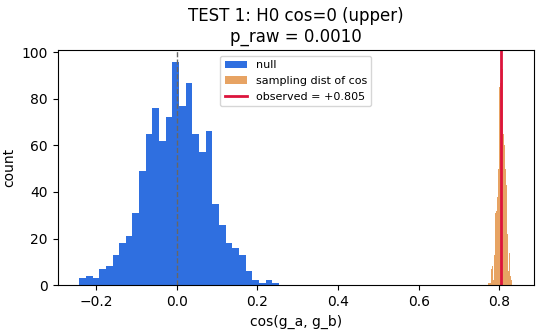}
  \caption{cos=0 test distribution for the Arousal dimension of Polish - English pair.}
\end{figure}

\begin{figure}[H]
  \includegraphics[width=\columnwidth]{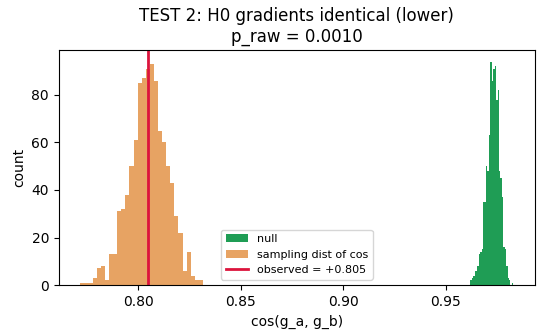}
  \caption{cos=1 test distribution for the Arousal dimension of Polish - English pair.}
\end{figure}

\subsubsection{Dominance}

\begin{figure}[H]
  \includegraphics[width=\columnwidth]{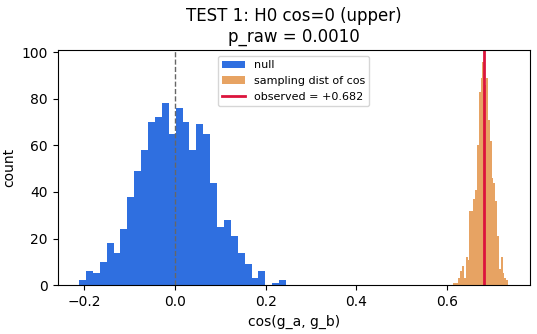}
  \caption{cos=0 test distribution for the Dominance dimension of Polish - English pair.}
\end{figure}

\begin{figure}[H]
  \includegraphics[width=\columnwidth]{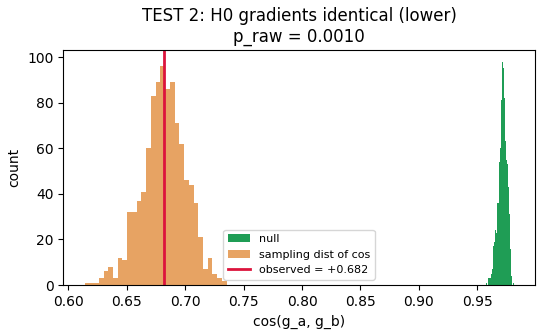}
  \caption{cos=1 test distribution for the Dominance dimension of Polish - English pair.}
\end{figure}

\subsection{Polish -- French}

\subsubsection{Valence}

\begin{figure}[H]
  \includegraphics[width=\columnwidth]{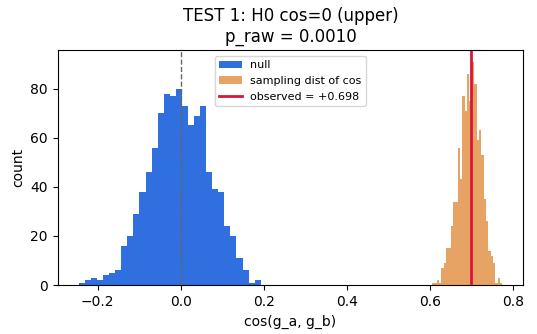}
  \caption{cos=0 test distribution for the Valence dimension of Polish - French pair.}
\end{figure}

\begin{figure}[H]
  \includegraphics[width=\columnwidth]{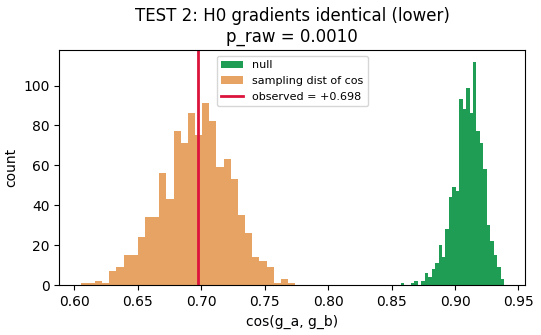}
  \caption{cos=1 test distribution for the Valence dimension of Polish - French pair.}
\end{figure}

\subsubsection{Arousal}

\begin{figure}[H]
  \includegraphics[width=\columnwidth]{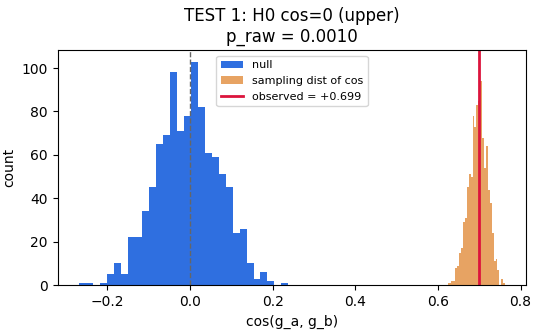}
  \caption{cos=0 test distribution for the Arousal dimension of Polish - French pair.}
\end{figure}

\begin{figure}[H]
  \includegraphics[width=\columnwidth]{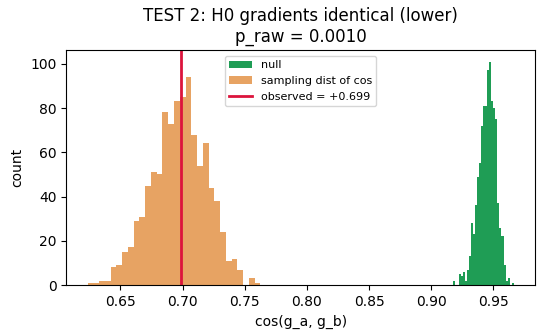}
  \caption{cos=1 test distribution for the Arousal dimension of Polish - French pair.}
\end{figure}

\subsection{English -- French}

\subsubsection{Valence}

\begin{figure}[H]
  \includegraphics[width=\columnwidth]{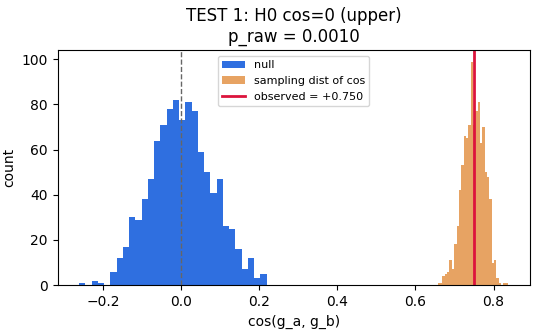}
  \caption{cos=0 test distribution for the Valence dimension of English - French pair.}
\end{figure}

\begin{figure}[H]
  \includegraphics[width=\columnwidth]{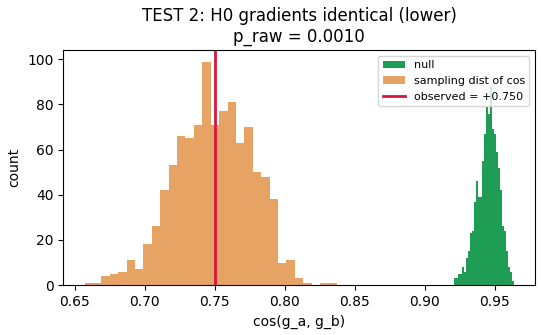}
  \caption{cos=1 test distribution for the Valence dimension of English - French pair.}
\end{figure}

\subsubsection{Arousal}

\begin{figure}[H]
  \includegraphics[width=\columnwidth]{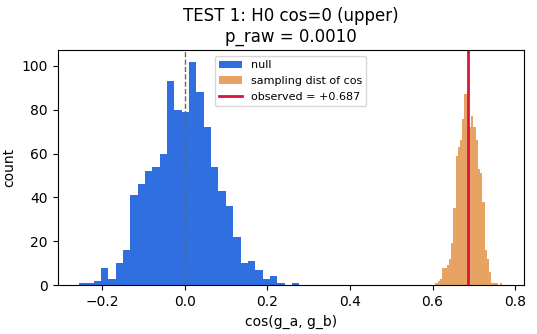}
  \caption{cos=0 test distribution for the Arousal dimension of English - French pair.}
\end{figure}

\begin{figure}[H]
  \includegraphics[width=\columnwidth]{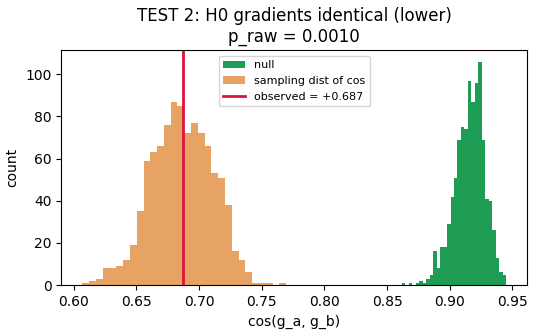}
  \caption{cos=1 test distribution for the Arousal dimension of English - French pair.}
\end{figure}

\section{Difference Gradient Clusters Report}
\label{sec:full-clusters}
In this appendix, we include the full cluster report, which is the result of k-means silhouette clustering around $\Delta{\vec{g}}$ and $-\Delta\vec{g}$ gradients. For every cluster, $n$ indicating the number of words in the cluster, $cos$ indicating the cosine similarity between the cluster centroid and the $\Delta{\vec{g}}$ gradient, and $coh.$ indicating the internal coherence (mean of cosine similarity between all pairs of cluster elements) of the cluster are provided. The words were translated from the original languages to English for the sake of presentation.

\subsection{English -- Polish Difference}

\subsubsection{Valence}

\subsubsection*{\underline{Neighbours in Polish Vocabulary}}

\noindent\textbf{Positive Pole (Leans Polish)}
\begin{itemize}
    \item \textbf{Cluster 0} ($n=28$, $\cos=0.41$, $\text{coh.}=0.45$): of Opole, of Tarnów, of Kraków, of Lesser Poland, of Wrocław, Tarnów, Opole-region, Sącz, Radlin, Tarnów-related.
    \item \textbf{Cluster 1} ($n=25$, $\cos=0.43$, $\text{coh.}=0.41$): modern, practical, contemporary/former, poetic, creative, medieval, old/former, living/vivid, pedagogical, unusual.
    \item \textbf{Cluster 2} ($n=47$, $\cos=0.54$, $\text{coh.}=0.34$): talented, talented people, performed/wrote, valued, presented, the youngest, reached for, conducting/leading, educated, became.
\end{itemize}

\noindent\textbf{Negative Pole (Leans English)}
\begin{itemize}
    \item \textbf{Cluster 0} ($n=30$, $\cos=-0.37$, $\text{coh.}=0.58$): accusation, accusations, suspicions, accused people, accusations, the accused person, suspicion, charges, with a charge, with charges.
    \item \textbf{Cluster 1} ($n=32$, $\cos=-0.45$, $\text{coh.}=0.49$): blocking, to block, to cause, to prevent, stopping, interruption, blocked, prevention, blocking, immediate.
    \item \textbf{Cluster 2} ($n=26$, $\cos=-0.37$, $\text{coh.}=0.56$): statement, confirmations, statements, as confirmation, confirmation, proven, confirmed, to confirm, as a statement, confirmed.
    \item \textbf{Cluster 3} ($n=12$, $\cos=-0.32$, $\text{coh.}=0.67$): cause, consequence, result, effect, consequences, causes, cause, in consequence, to contribute, consequences.
\end{itemize}

\subsubsection*{\underline{Neighbours in English Vocabulary}}

\noindent\textbf{Positive Pole (Leans Polish)}
\begin{itemize}
    \item \textbf{Cluster 0} ($n=52$, $\cos=0.33$, $\text{coh.}=0.56$): pretentious, arrogant, overbearing, pompous, obnoxious, inept, stereotypical, unattractive, uneducated, stereotyped.
    \item \textbf{Cluster 1} ($n=48$, $\cos=0.57$, $\text{coh.}=0.31$): vertically, tapered, strung, shifting, picker, oriented, spins, drag, immersed, cylindrical.
\end{itemize}

\noindent\textbf{Negative Pole (Leans English)}
\begin{itemize}
    \item \textbf{Cluster 0} ($n=43$, $\cos=-0.36$, $\text{coh.}=0.58$): substantiated, conclusive, evidence, substantiate, corroborate, confirm, conclusively, corroborated, circumstantial, contradicted.
    \item \textbf{Cluster 1} ($n=38$, $\cos=-0.44$, $\text{coh.}=0.46$): caused, prevented, cause, damage, resulted, occurred, prevent, precaution, triggered, accidental.
    \item \textbf{Cluster 2} ($n=19$, $\cos=-0.32$, $\text{coh.}=0.61$): unblocked, indeffed, blocked, unblocking, unblock, indefinitely, muzemike, –muzemike, evading, disruption.
\end{itemize}

\subsubsection{Arousal}

\subsubsection*{\underline{Neighbours in Polish Vocabulary}}

\noindent\textbf{Positive Pole (Leans Polish)}
\begin{itemize}
    \item \textbf{Cluster 0} ($n=23$, $\cos=0.45$, $\text{coh.}=0.53$): killed, shot/wounded by gunfire, kills, fatally, shot dead, shot dead/wounded, killing, wounding, with a knife, police officer.
    \item \textbf{Cluster 1} ($n=24$, $\cos=0.40$, $\text{coh.}=0.61$): suffered from illness, tuberculosis, fell ill, sick/ill, illness, heart attack, suffered, died, depression, Parkinson's disease.
    \item \textbf{Cluster 2} ($n=21$, $\cos=0.41$, $\text{coh.}=0.57$): stopped/detained, attacked, murdered, detained, deprived, committed/allowed, fearing, killed, feared, attacked.
    \item \textbf{Cluster 3} ($n=16$, $\cos=0.39$, $\text{coh.}=0.62$): commission, to commit, committed, suicide, committed, commits, suicide, committed, massacre, the accused person.
    \item \textbf{Cluster 4} ($n=16$, $\cos=0.41$, $\text{coh.}=0.65$): caused, caused by, caused by, caused by, caused by, caused, caused, caused, induced by, causing.
\end{itemize}

\noindent\textbf{Negative Pole (Leans English)}
\begin{itemize}
    \item \textbf{Cluster 0} ($n=37$, $\cos=-0.42$, $\text{coh.}=0.53$): artistic, artistic-, multimedia, architectural, artistic, artistic, visual arts, artistic, photographic, educational.
    \item \textbf{Cluster 1} ($n=18$, $\cos=-0.50$, $\text{coh.}=0.44$): colorful, ornamental, decorative, ornamentation, color scheme, exotic, floral, exotic, shapes, flower.
    \item \textbf{Cluster 2} ($n=21$, $\cos=-0.50$, $\text{coh.}=0.43$): astronomy, mathematical, astronomical, mathematical, mathematics, astronomical, gymnastics, gymnastic, geometric, canoeing/kayaking.
    \item \textbf{Cluster 3} ($n=24$, $\cos=-0.36$, $\text{coh.}=0.61$): attractions, tourist, tourist, attraction, attractions, tourist, attractive, scenic/viewpoint-related, qualities/assets, sightseeing-related.
\end{itemize}

\subsubsection*{\underline{Neighbours in English Vocabulary}}

\noindent\textbf{Positive Pole (Leans Polish)}
\begin{itemize}
    \item \textbf{Cluster 0} ($n=58$, $\cos=0.47$, $\text{coh.}=0.55$): crippling, exacerbated, blamed, worsened, worsening, blaming, suffer, suffered, succumbing, caused.
    \item \textbf{Cluster 1} ($n=42$, $\cos=0.42$, $\text{coh.}=0.63$): debilitating, chronic, illness, hospitalization, ailment, hospitalized, paralysis, symptomatic, suffering, degenerative.
\end{itemize}

\noindent\textbf{Negative Pole (Leans English)}
\begin{itemize}
    \item \textbf{Cluster 0} ($n=35$, $\cos=-0.44$, $\text{coh.}=0.46$): ecotourism, tourism, touristic, sightseeing, birding, kayaking, excursions, rafting, archeology, scenic.
    \item \textbf{Cluster 1} ($n=65$, $\cos=-0.47$, $\text{coh.}=0.43$): showcasing, showcases, colorful, vibrant, showcase, fascinating, exciting, colourful, unique, elegant.
\end{itemize}

\subsubsection{Dominance}

\subsubsection*{\underline{Neighbours in Polish Vocabulary}}

\noindent\textbf{Positive Pole (Leans Polish)}
\begin{itemize}
    \item \textbf{Cluster 0} ($n=54$, $\cos=0.47$, $\text{coh.}=0.50$): terrorist, terrorist-related, terrorist-related, terrorists, attacks, terrorist, attacks, militias, terrorism, terrorism.
    \item \textbf{Cluster 1} ($n=46$, $\cos=0.52$, $\text{coh.}=0.44$): rulers, ruler, rulers, monarch, dynasty, dynasty, monarchs, monarchy, governor/viceroy, most powerful.
\end{itemize}

\noindent\textbf{Negative Pole (Leans English)}
\begin{itemize}
    \item \textbf{Cluster 0} ($n=50$, $\cos=-0.57$, $\text{coh.}=0.35$): to sleep, boys/guys, breakfast, sleeps, spend time, beds, to spend, vacation/holidays, to drink, moments.
    \item \textbf{Cluster 1} ($n=50$, $\cos=-0.47$, $\text{coh.}=0.42$): submission/report, with a submission/report, draft, to be sufficient, I submitted/reported, is sufficient, will harm, correction/improvement, draft, I completed/supplemented.
\end{itemize}

\subsubsection*{\underline{Neighbours in English Vocabulary}}

\noindent\textbf{Positive Pole (Leans Polish)}
\begin{itemize}
    \item \textbf{Cluster 0} ($n=23$, $\cos=0.42$, $\text{coh.}=0.56$): terrorist, terrorists, terrorism, assassinations, gangs, cartels, insurgency, robberies, vigilante, terror.
    \item \textbf{Cluster 1} ($n=17$, $\cos=0.40$, $\text{coh.}=0.57$): weapon, weapons, weaponry, warheads, thermonuclear, nuclear, wields, wielding, wielded, atomic.
    \item \textbf{Cluster 2} ($n=27$, $\cos=0.42$, $\text{coh.}=0.54$): warlords, overthrowing, tyrannical, rulers, militarily, ruler, overthrow, empires, warlord, imperialist.
    \item \textbf{Cluster 3} ($n=33$, $\cos=0.43$, $\text{coh.}=0.54$): demonic, warlock, evil, juggernaut, monstrous, demon, sorcerer, fearsome, scourge, monster.
\end{itemize}

\noindent\textbf{Negative Pole (Leans English)}
\begin{itemize}
    \item \textbf{Cluster 0} ($n=52$, $\cos=-0.47$, $\text{coh.}=0.46$): meals, kitchen, toilets, meal, toilet, pantry, bathrooms, bathroom, cooked, washing.
    \item \textbf{Cluster 1} ($n=48$, $\cos=-0.44$, $\text{coh.}=0.47$): hopefully, glad, tidying, apologies, userfication, reconsider, userfy, revisit, decent, userfied.
\end{itemize}

\subsection{Polish -- French Difference}
\subsubsection{Valence}

\subsubsection*{\underline{Neighbours in Polish Vocabulary}}

\noindent\textbf{Positive Pole (Leans Polish)}
\begin{itemize}
    \item \textbf{Cluster 0} ($n=38$, $\cos=0.46$, $\text{coh.}=0.40$): raw material, stamens, pistil, floral, stamens, pigment/dye, pistils, plant-based, plant, ornamental.
    \item \textbf{Cluster 1} ($n=62$, $\cos=0.40$, $\text{coh.}=0.46$): Ostrowiec, Łagiewniki, Psary, Piekary, Ostrowy, brickworks, Ostrówek, Ostrów, Trzebinia, Wolsztyn.
\end{itemize}

\noindent\textbf{Negative Pole (Leans French)}
\begin{itemize}
    \item \textbf{Cluster 0} ($n=46$, $\cos=-0.45$, $\text{coh.}=0.51$): accusation, suspicions, suspicion, accusations, violations, violation, charges, with a charge, with charges, disclosure.
    \item \textbf{Cluster 1} ($n=23$, $\cos=-0.34$, $\text{coh.}=0.68$): killings/homicides, murder, killing/homicide, murder, murders, murdered, killing, killing, killed, in a killing/homicide.
    \item \textbf{Cluster 2} ($n=31$, $\cos=-0.51$, $\text{coh.}=0.44$): fearing, agreed, to accept, feared, agreed, demanding, outrage, to deprive, threatened, deprived.
\end{itemize}

\subsubsection*{\underline{Neighbours in French Vocabulary}}

\noindent\textbf{Positive Pole (Leans Polish)}
\begin{itemize}
    \item \textbf{Cluster 0} ($n=26$, $\cos=0.33$, $\text{coh.}=0.53$): anode, solvent, cathode, electrode, emulsion, porous, solvents, microscopic, microscopic, microorganisms.
    \item \textbf{Cluster 1} ($n=23$, $\cos=0.42$, $\text{coh.}=0.41$): mechanization, industrially, fertilizers, industrial, ore, coal, mechanized, Zamość, Gorzów, extraction.
    \item \textbf{Cluster 2} ($n=21$, $\cos=0.39$, $\text{coh.}=0.49$): cylinder, crank, valve, container/vessel, belt, pneumatic, millstone/grindstone, funnel, adjustment, rope/cordage.
    \item \textbf{Cluster 3} ($n=20$, $\cos=0.34$, $\text{coh.}=0.52$): rectangular, cylindrical, rectangular, polygonal, cylindrical, square, square, diameter, support/structural support, round.
    \item \textbf{Cluster 4} ($n=10$, $\cos=0.33$, $\text{coh.}=0.51$): supplement, expedition, warfare, petroleum, Pacific, monitor, \textit{Trichomanes}, \textit{intermedia}, hundred, Kuntze.
\end{itemize}

\noindent\textbf{Negative Pole (Leans French)}
\begin{itemize}
    \item \textbf{Cluster 0} ($n=53$, $\cos=-0.61$, $\text{coh.}=0.34$): Proulx, Tremblay, Gagnon, Gélinas, Beaudoin, Trudel, Falardeau, Marois, Dubé, Dionne.
    \item \textbf{Cluster 1} ($n=47$, $\cos=-0.44$, $\text{coh.}=0.48$): dies, passes away, to pass away, remarriage, to remarry, passed away, died, occurrence, gives birth, marriage.
\end{itemize}

\subsubsection{Arousal}

\subsubsection*{\underline{Neighbours in Polish Vocabulary}}

\noindent\textbf{Positive Pole (Leans Polish)}
\begin{itemize}
    \item \textbf{Cluster 0} ($n=56$, $\cos=0.58$, $\text{coh.}=0.43$): damage/injury, damages/injuries, swelling/edema, paralysis, bleeding, bronchi, damage/injury, ruptures/cracks, narrowing/stenosis, chronic.
    \item \textbf{Cluster 1} ($n=44$, $\cos=0.46$, $\text{coh.}=0.53$): murdered, was/were murdered, execution by shooting, murders, killed, executed by shooting, committed, perpetrators, killed, shot/executed.
\end{itemize}

\noindent\textbf{Negative Pole (Leans French)}
\begin{itemize}
    \item \textbf{Cluster 0} ($n=62$, $\cos=-0.56$, $\text{coh.}=0.41$): desires, desire, perfection, spiritual, joy, happiness, beauty, dreams, beauty, understanding.
    \item \textbf{Cluster 1} ($n=38$, $\cos=-0.56$, $\text{coh.}=0.42$): landscapes, travels, nature, trip/excursion, landscapes, nature, traveler, journey/travel, exotic, sightseeing-related.
\end{itemize}

\subsubsection*{\underline{Neighbours in French Vocabulary}}

\noindent\textbf{Positive Pole (Leans Polish)}
\begin{itemize}
    \item \textbf{Cluster 0} ($n=53$, $\cos=0.61$, $\text{coh.}=0.37$): barricade, landslide/collapse, assailants, entrenchments, structural support, collapse, collapses, derailment, obstruction/chicane, ditch.
    \item \textbf{Cluster 1} ($n=47$, $\cos=0.41$, $\text{coh.}=0.55$): lesion/injury, hernia, hip, lesions/injuries, joint-related, fracture, knee, tendon, hemorrhage, fractures.
\end{itemize}

\noindent\textbf{Negative Pole (Leans French)}
\begin{itemize}
    \item \textbf{Cluster 0} ($n=23$, $\cos=-0.40$, $\text{coh.}=0.66$): desserts, pastries, cakes, sweet, to taste/enjoy, ingredients, culinary, confectionery, flavors, culinary.
    \item \textbf{Cluster 1} ($n=32$, $\cos=-0.56$, $\text{coh.}=0.46$): folkloric, folkloric, playful, poetic, music, creative, folklore, traditions, exoticism, playful.
    \item \textbf{Cluster 2} ($n=45$, $\cos=-0.50$, $\text{coh.}=0.52$): kindness, joy, tenderness, benevolence, generosity, goodness, gratitude, serenity, generous, gentleness/sweetness.
\end{itemize}

\subsection{English -- French Difference}

\subsubsection{Valence}

\subsubsection*{\underline{Neighbours in English Vocabulary}}

\noindent\textbf{Positive Pole (Leans English)}
\begin{itemize}
    \item \textbf{Cluster 0} ($n=20$, $\cos=0.48$, $\text{coh.}=0.38$): mill, grist, sawmill, storehouse, powder, account, hearth, accounts, tools, loaf.
    \item \textbf{Cluster 1} ($n=26$, $\cos=0.35$, $\text{coh.}=0.51$): juniata, middletown, kanawha, muskingum, hunterdon, raritan, allegheny, shenandoah, merrimack, brandywine.
    \item \textbf{Cluster 2} ($n=18$, $\cos=0.33$, $\text{coh.}=0.53$): neamț, hunedoara, buzău, valea, wysokie, sibiu, stara, baia, kardzhali, dąbrowa.
    \item \textbf{Cluster 3} ($n=16$, $\cos=0.39$, $\text{coh.}=0.46$): unlocking, unlocked, kindle, reset, battery, nexus, pocket, finder, button, tablet.
    \item \textbf{Cluster 4} ($n=20$, $\cos=0.31$, $\text{coh.}=0.57$): potatoes, potato, maize, tubers, cassava, flour, flax, crops, seeds, carrot.
\end{itemize}

\noindent\textbf{Negative Pole (Leans French)}
\begin{itemize}
    \item \textbf{Cluster 0} ($n=17$, $\cos=-0.43$, $\text{coh.}=0.50$): vous, sont, chez, espace, cinéma, gainsbourg, québécois, tremblay, dit, artiste.
    \item \textbf{Cluster 1} ($n=83$, $\cos=-0.43$, $\text{coh.}=0.50$): overbearing, arrogant, obnoxious, pretentious, pompous, inept, condescending, demeaning, unattractive, unhappy.
\end{itemize}

\subsubsection*{\underline{Neighbours in French Vocabulary}}

\noindent\textbf{Positive Pole (Leans English)}
\begin{itemize}
    \item \textbf{Cluster 0} ($n=11$, $\cos=0.25$, $\text{coh.}=0.73$): Jablanica, Moravica, Pomoravlje, Nišava, Toplica, Pljevlja, Donja, Stara, Cyrillic, iconostasis.
    \item \textbf{Cluster 1} ($n=46$, $\cos=0.49$, $\text{coh.}=0.37$): locking, ammunition, lock/bolt, to activate, activated, activated, mechanism, hold/storage bay, blocking, casemate.
    \item \textbf{Cluster 2} ($n=43$, $\cos=0.31$, $\text{coh.}=0.61$): bacteria, bacterial, bacterial, bacterial, microorganisms, bacterium, enzymes, antibiotic, antibiotics, carbohydrates.
\end{itemize}

\noindent\textbf{Negative Pole (Leans French)}
\begin{itemize}
    \item \textbf{Cluster 0} ($n=33$, $\cos=-0.40$, $\text{coh.}=0.51$): Lévesque, Gagnon, Tremblay, Beaudoin, Séguin, Bélanger, Simard, Québécois, Falardeau, Gélinas.
    \item \textbf{Cluster 1} ($n=67$, $\cos=-0.53$, $\text{coh.}=0.38$): marry, to marry, to remarry, move house, remarriage, marriage, remarried, invite, neglects/abandons, aunts.
\end{itemize}

\subsubsection{Arousal}

\subsubsection*{\underline{Neighbours in English Vocabulary}}

\noindent\textbf{Positive Pole (Leans English)}
\begin{itemize}
    \item \textbf{Cluster 0} ($n=35$, $\cos=0.50$, $\text{coh.}=0.41$): vaginal, penis, vagina, genital, ejaculation, nipple, anus, groin, incision, anal.
    \item \textbf{Cluster 1} ($n=65$, $\cos=0.42$, $\text{coh.}=0.50$): cylindrical, angled, conical, tapered, protruding, diameters, curved, perpendicular, concentric, casing.
\end{itemize}

\noindent\textbf{Negative Pole (Leans French)}
\begin{itemize}
    \item \textbf{Cluster 0} ($n=27$, $\cos=-0.45$, $\text{coh.}=0.59$): longing, yearning, sorrow, despair, sadness, loneliness, sufferings, dreams, laments, joys.
    \item \textbf{Cluster 1} ($n=42$, $\cos=-0.45$, $\text{coh.}=0.57$): affection, unhappiness, unhappy, saddened, distrust, kindness, dismayed, disillusioned, compassion, overjoyed.
    \item \textbf{Cluster 2} ($n=13$, $\cos=-0.40$, $\text{coh.}=0.65$): recounts, recounted, recollection, experiences, recollections, narrates, remembers, memories, narrator, experience.
    \item \textbf{Cluster 3} ($n=18$, $\cos=-0.37$, $\text{coh.}=0.71$): father, mother, grandmother, grandfather, grandparents, parents, siblings, stepmother, uncle, adoptive.
\end{itemize}

\subsubsection*{\underline{Neighbours in French Vocabulary}}

\noindent\textbf{Positive Pole (Leans English)}
\begin{itemize}
    \item \textbf{Cluster 0} ($n=53$, $\cos=0.45$, $\text{coh.}=0.48$): perpendicularly, perpendicular, friction, curvature, torsion, perpendicular, winding/coiling, laterally, friction, vertically.
    \item \textbf{Cluster 1} ($n=47$, $\cos=0.43$, $\text{coh.}=0.52$): corbeling, moldings, polygonal, machicolations, molding, entablature, cylindrical, buttresses, projecting/protruding, vaults.
\end{itemize}

\noindent\textbf{Negative Pole (Leans French)}
\begin{itemize}
    \item \textbf{Cluster 0} ($n=17$, $\cos=-0.40$, $\text{coh.}=0.63$): feared, recalled, explained, felt, feared, fearing, expressed, believed, claiming as a pretext, expressed.
    \item \textbf{Cluster 1} ($n=47$, $\cos=-0.47$, $\text{coh.}=0.55$): sadness, resentment, despair, feeling, feelings, distress/disarray, joy, regret, worry/anxiety, feels/experiences.
    \item \textbf{Cluster 2} ($n=14$, $\cos=-0.43$, $\text{coh.}=0.61$): father, uncle, uncles, brother, paternal, parents, mother, elders/older siblings, friend, native/birthplace.
    \item \textbf{Cluster 3} ($n=22$, $\cos=-0.47$, $\text{coh.}=0.56$): recounts/tells, recounted/told, recounting/telling, lived/experienced, recounted/told, relates, recalls/remembers, lived/experienced, childhood, remembers.
\end{itemize}

\end{document}